\documentclass{article}

\usepackage{amsfonts}
\usepackage{amsopn}
\usepackage{amsmath}
\usepackage{amssymb}
\usepackage{mathtools}
\usepackage{bbm}
\usepackage{color}
\usepackage{amsthm}
\usepackage{placeins}
\usepackage{wrapfig}
\usepackage{float}
\usepackage{nicefrac}

\makeatletter  
\let\NAT@parse\undefined
\makeatother

\usepackage{hyperref}

\newcommand{\eg}{e.g.\ }

\newcommand{\E}[2]{\operatorname{\mathbb{E}}_{#1}\left[#2\right]}

\newcommand{\density}{p}

\newcommand{\kl}[2]{\mathrm{D_{KL}}\left(#1\;\middle\|\;#2\right)}

\newcommand{\ent}{\mathcal{H}}

\newcommand{\voidarg}{{\,\cdot\,}}

\newcommand{\sspace}{\mathcal{S}}
\newcommand{\aspace}{\mathcal{A}}

\newcommand{\state}{\mathbf{s}}
\newcommand{\sz}{{\state_0}}

\newcommand{\st}{{\state_t}}

\newcommand{\stp}{{\state_{t+1}}}

\newcommand{\pdyn}{\density_\state}

\newcommand{\action}{\mathbf{a}}

\newcommand{\at}{{\action_t}}

\newcommand{\opt}{^*}

\newcommand{\latentt}{\mathbf{h}_t}

\newcommand{\optim}{\mathcal{O}}
\newcommand{\optimt}{\optim_t}

\newcommand{\reward}{r}

\newcommand{\rt}{\reward_t}

\newcommand{\rmin}{r_\mathrm{min}}
\newcommand{\rmax}{r_\mathrm{max}}

\newcommand{\policy}{\pi}

\newcommand{\inv}{^{-1}}

\newcommand{\reals}{\mathbb{R}}

\newcommand{\aref}[1]{\hyperref[#1]{Appendix~\ref*{#1}}}

\def\alignautorefname~#1\null{%
  (#1)\null
}\def\equationautorefname~#1\null{%
  Equation~#1\null
}

\usepackage{microtype}
\usepackage{graphicx}
\usepackage{subfigure}
\usepackage{booktabs} %

\usepackage{hyperref}

\usepackage[accepted]{icml2018}

\icmltitlerunning{Latent Space Policies for Hierarchical Reinforcement Learning}

\begin{document}

\twocolumn[
\icmltitle{Latent  Space Policies for Hierarchical Reinforcement Learning}
\icmlsetsymbol{equal}{*}

\begin{icmlauthorlist}
\icmlauthor{Tuomas Haarnoja}{equal,berkeley}
\icmlauthor{Kristian Hartikainen}{equal,ind}
\icmlauthor{Pieter Abbeel}{berkeley}
\icmlauthor{Sergey Levine}{berkeley}
\end{icmlauthorlist}

\icmlaffiliation{berkeley}{Berkeley Artificial Intelligence Research, University of California, Berkeley, USA}
\icmlaffiliation{ind}{Independent researcher, Seattle, WA, USA}

\icmlcorrespondingauthor{Tuomas Haarnoja}{haarnoja@berkeley.edu}
\icmlcorrespondingauthor{Kristian Hartikainen}{kristian.hartikainen@gmail.com}

\icmlkeywords{reinforcement learning, hierarchical reinforcement learning, control as inference}

\vskip 0.3in
]

\printAffiliationsAndNotice{\icmlEqualContribution} %

\begin{abstract}
We address the problem of learning hierarchical deep neural network policies for reinforcement learning. In contrast to methods that explicitly restrict or cripple lower layers of a hierarchy to force them to use higher-level modulating signals, each layer in our framework is trained to directly solve the task, but acquires a range of diverse strategies via a maximum entropy reinforcement learning objective. Each layer is also augmented with latent random variables, which are sampled from a prior distribution during the training of that layer. The maximum entropy objective causes these latent variables to be incorporated into the layer's policy, and the higher level layer can directly control the behavior of the lower layer through this latent space. Furthermore, by constraining the mapping from latent variables to actions to be invertible, higher layers retain full expressivity: neither the higher layers nor the lower layers are constrained in their behavior. Our experimental evaluation demonstrates that we can improve on the performance of single-layer policies on standard benchmark tasks simply by adding additional layers, and that our method can solve more complex sparse-reward tasks by learning higher-level policies on top of high-entropy skills optimized for simple low-level objectives.
\end{abstract}

\vspace{-2mm}
\section{Introduction}
\vspace{-1mm}

Model-free deep reinforcement learning (RL) has demonstrated potential in many challenging domains, ranging from games \citep{mnih2013playing,silver2016mastering} to manipulation \citep{levine2016end} and locomotion task \cite{schulman2015trust}. Part of the promise of incorporating deep representations into RL is the potential for the emergence of hierarchies, which can enable reasoning and decision making at different levels of abstract. A hierarchical RL algorithm could, in principle, efficiently discover solutions to complex problems and reuse representations between related tasks. While hierarchical structures have been observed to emerge in deep networks applied to perception tasks, such as computer vision~\cite{lbh-dl-15}, it remains an open question how suitable hierarchical representations can be induced in a reinforcement learning setting. A central challenge with these methods is the automation of the hierarchy construction process: hand-specified hierarchies require considerable expertise and insight to design and limit the generality of the approach \citep{sutton1999between,kulkarni2016hierarchical,tessler2017deep}, while automated methods must contend with severe challenges, such as the collapse of all primitives into just one useful skill \citep{bacon2017option} or the need to hand-engineer primitive discovery objectives or intermediate goals \citep{heess2016learning}.

When learning hierarchies automatically, we must answer a critical question: What objective can we use to ensure that lower layers in a hierarchy are useful to the higher layers? Prior work has proposed a number of heuristic approaches, such as hiding some parts of the observation from the lower layers~\citep{heess2016learning}, hand-designing state features and training the lower layer behaviors to maximize mutual information against these features~\citep{florensa2017stochastic}, or constructing diversity-seeking priors that cause lower-layer primitives to take on different roles~\citep{daniel2012hierarchical,eysenbach2018diversity}. Oftentimes, these heuristics intentionally cripple the lower layers of the hierarchy, for example, by withholding information, so as to force a hierarchy to emerge, or else limit the higher levels of the hierarchy to selecting from among a discrete set of skills~\citep{bacon2017option}. In both cases, the hierarchy is forced to emerge because neither higher nor lower layers can solve the problem alone. However, constraining the layers in this way can involve artificial and task-specific restrictions~\cite{heess2016learning} or else diminish overall performance.

Instead of crippling or limiting the different levels of the hierarchy, we can imagine a hierarchical framework in which each layer directly attempts to solve the task and, if it is not fully successful, makes the job easier for the layer above it. In this paper, we explore a solution to the hierarchical reinforcement learning problem based on this principle. In our framework, each layer of the hierarchy corresponds to a policy with internal latent variables. These latent variables determine how the policy maps states into actions, and the latent variables of the lower-level policy act as the action space for the higher level. Crucially, each layer is unconstrained, both in its ability to sense and affect the environment: each layer receives the full state as the observation, and each layer is, by construction, fully invertible, so that higher layers in the hierarchy can undo any transformation of the action space imposed on the layers below.

In order to train policies with latent variables, we cast the problem of reinforcement learning into the framework of probabilistic graphical models. To that end, we build on maximum entropy reinforcement learning~\citep{todorov2006linearly,ziebart2008maximum}, where the RL objective is modified to optimize for stochastic policies that maximize both reward and entropy. It can be shown that, in this framework, the RL problem becomes equivalent to an inference problem in a particular type of probabilistic graphical model~\cite{toussaint2009robot}. By augmenting this model with latent variables, we can derive a method that simultaneously produces a policy that attempts to solve the task, and a latent space that can be used by a higher-level controller to \emph{steer} the policy's behavior.

The particular latent variable model representation that we use is based on normalizing flows~\citep{dinh2016density} that transform samples from a spherical Gaussian prior latent variable distribution into a posterior distribution, which in the case of our policies corresponds to a distribution over actions. When this transformation is described by a general-purpose neural network, the model can represent any distribution over the observed variable when the network is large enough. By conditioning the entire generation process on the state, we obtain a policy that can represent any conditional distribution over actions. When combined with maximum entropy reinforcement learning algorithms, this leads to a RL method that is expressive, powerful, and surprisingly stable. In fact, our experimental evaluation shows that this approach can attain state-of-the-art results on a number of continuous control benchmark tasks by itself, independently of its applicability to hierarchical RL. 

The contributions of our paper consist of a stable and scalable algorithm for training maximum entropy policies with latent variables as well as a framework for constructing hierarchies out of these latent variable policies. Hierarchies are constructed in layerwise fashion, by training one latent variable policy at a time, with each policy using the latent space of the policy below it as an action space, as illustrated in \autoref{fig:network}. Each layer can be trained either on the true reward for the task, without any modification, or on a lower-level shaping reward. For example, for learning a complex navigation task, lower layers might receive a reward that promotes locomotion, regardless of direction, while higher layers aim to reach a particular location. 
When the shaping terms are not available, the same reward function can be used for each layer, and we still observe significant improvements from hierarchy. Our experimental evaluation illustrates that our method produces state-of-the-art results in terms of sample complexity on a variety of benchmark tasks, including a humanoid robot with 21 actuators, even when training a single layer of the hierarchy, and can further improve performance when additional layers are added. Furthermore, we illustrate that more challenging tasks with sparse reward signals can be solved effectively by providing shaping rewards to lower layers. 

\vspace{-1mm}
\section{Related Work}
\label{sec:related}
\vspace{-1mm}

A number of prior works have explored how reinforcement learning can be cast in the framework of probabilistic inference \citep{kappen2005path,todorov2006linearly,ziebart2008maximum,toussaint2009robot,peters2010relative,neumann2011variational}. Our approach is based on a formulation of reinforcement learning as inference in a graphical model~\cite{ziebart2008maximum,toussaint2009robot,levine2014motor}. %
Prior work has shown that this framework leads to an entropy-maximizing version of reinforcement learning, where the standard objective is augmented with a term that also causes the policy to maximize entropy~\cite{ziebart2008maximum,haarnoja2017reinforcement,haarnoja2018soft,nachum2017bridging,schulman2017equivalence}. Intuitively, this encourages policies that maximize reward while also being as random as possible, which can be useful for robustness, exploration, and, in our case, increasing the diversity of behaviors for lower layers in a hierarchy. Building on this graphical model interpretation of reinforcement learning also makes it natural for us to augment the policy with latent variables. While several prior works have sought to combine maximum entropy policies with learning of latent spaces~\cite{haarnoja2017reinforcement,hausman2018learning} and even with learning hierarchies in small state spaces~\cite{saxe2017hierarchy}, to our knowledge, our method is the first to extend this mechanism to the setting of learning hierarchical policies with deep RL in continuous domains. %

Prior frameworks for hierarchical learning are often based on either options or contextual policies. The options framework \citep{sutton1999between} combines low-level option policies with a top-level policy that invokes individual options, whereas contextual policies \citep{kupcsik2013data,schaul2015universal,heess2016learning} generalize options to continuous goals. One of the open questions in both options and contextual policy frameworks is how the base policies should be acquired. In some situations, a reasonable solution is to resort to domain knowledge and design a span of subgoals manually \cite{heess2016learning,kulkarni2016hierarchical,macalpine2018overlapping}. Another option is to train the entire hierarchy end-to-end~\citep{bacon2017option,vezhnevets2017feudal,daniel2012hierarchical}. While the end-to-end training scheme provides generality and flexibility, it is prone to learning degenerate policies that exclusively use a single option, losing much of the benefit of the hierarchical structure \citep{bacon2017option}. To that end, the option-critic \citep{bacon2017option} adopts a standard entropy regularization scheme ubiquitous in policy gradient methods \citep{mnih2016asynchronous,schulman2015trust}, \citeauthor{florensa2017stochastic} propose maximizing the mutual information of the top-level actions and the state distribution, and \citeauthor{daniel2012hierarchical} bound the mutual information of the actions and top-level actions. Our method also uses entropy maximization to obtain diverse base policies, but in contrast to prior methods,
our sub-policies are invertible and parameterized by continuous latent variables. The higher levels can thus undo any lower level transformation, and the lower layers can learn independently, allowing us to train the hierarchies in bottom-up layerwise fashion. Unlike prior methods, which use structurally distinct higher and lower layers, all layers in our hierarchies are structurally generic, and are trained with exactly the same procedure.

\section{Preliminaries}
\label{sec:preliminaries}
In this section, we introduce notation and summarize standard and maximum entropy reinforcement learning.

\subsection{Notation}

We address policy learning in continuous action spaces formalized as learning in a Markov decision process (MDP) $(\sspace, \aspace, \pdyn, \reward)$,
where $\sspace$ and $\aspace$ represent the state and actions spaces, and $\pdyn:\ \sspace \times \sspace \times \aspace \rightarrow [0,\, \infty)$ represents the state transition probabilities of the next state $\stp\in\sspace$ given the current state $\st\in\sspace$ and action $\at\in\aspace$. At each transition, the environment emits a bounded reward $\reward: \sspace \times \aspace \rightarrow  [\rmin,\rmax]$. We will also use $p(\state_0)$ to denote the initial state distribution, $\tau = (\state_0, \action_0, ..., \state_T)$ to denote a trajectory and $\rho_\policy(\tau)$ its distribution under a policy $\policy(\at|\st)$.

\subsection{Maximum Entropy Reinforcement Learning}

The standard objective used in reinforcement learning is to maximize the expected sum of rewards $\sum_t \E{(\st,\at)\sim\rho_\policy}{\rt}$. We will consider a more general maximum entropy objective (see, \eg,~\citep{todorov2006linearly, ziebart2010modeling, rawlik2012stochastic, fox2015taming,haarnoja2017reinforcement,haarnoja2018soft}), which augments the objective with the expected entropy of the policy over $\rho_\policy(\tau)$:
\begin{align}
\label{eq:maxent_objective}
J(\policy)  =  \E{\tau\sim \rho_\policy(\tau)}{\sum_{t} \reward(\st,\at) + \alpha\ent(\policy(\voidarg|\st))}.
\end{align}
The temperature parameter $\alpha > 0$ determines the relative importance of the entropy term against the reward and thus controls the stochasticity of the optimal policy. 
The conventional objective can be recovered in the limit as $\alpha \rightarrow 0$. For the rest of this paper, we will omit writing the temperature explicitly, as it can always be subsumed into the reward by scaling it with $\alpha\inv$. In practice, we optimize a discounted, infinite horizon objective, which is more involved to write out explicitly, and we refer the interested readers to prior work for details~\citep{haarnoja2017reinforcement}.

\section{Control as Inference}

In this section, we derive the maximum entropy objective by transforming the optimal control problem
into an inference problem.
Our proposed hierarchical framework will later build off of this probabilistic view of optimal control (\autoref{sec:our_method}).

\subsection{Probabilistic Graphical Model for Control}
Our derivation is based on the probabilistic graphical model in \autoref{fig:pgm_no_policy}. 
This model is composed of factors for the dynamics $p(\stp|\st,\at)$ and for an action prior $p(\at)$, which is typically taken to be a uniform distribution but, as we will discuss later, will be convenient to set to a Gaussian distribution in the hierarchical case. Because we are interested in inferring the optimal trajectory distribution under a given reward function, we attach to each state and action a binary random variable $\optimt$, or \emph{optimality variable}, denoting whether the time step was ``optimal.'' To solve the optimal control problem, we can now infer the posterior action distribution $\policy\opt(\at | \st) = p(\at | \st, \optim_{t:T} = true)$, which simply states that an optimal action is such that the optimality variable is active for the current state and for all of the future states. For the remainder of this paper, we will refrain, for conciseness, from explicitly writing $\optimt=true$, and instead write $\optimt$ to denote the state-action tuple for the corresponding time was optimal.

A convenient way to incorporate reward function into this framework is to choose $p(\optimt|\st,\at) = \exp( \reward(\st,\at))$, assuming, without loss of generality, that $\reward(\st,\at) < 0$. 
We can write the distribution over optimal trajectories as
\begin{align}
\label{eq:optimal_trajectory_distribution}
p(\tau| \optim_{0:T}\!)\! &\propto p(\sz)\prod_{t=0}^{T} p(\at)p(\stp | \st, \at) \exp\!\left( \reward(\st, \at)\right)
\end{align}
and use it to make queries, such as $p(\at| \st, \optim_{t:T})$. As we will discuss in the next section, using variational inference to determine $p(\at| \st, \optim_{t:T})$ reduces to the familiar maximum entropy reinforcement learning problem in \autoref{eq:maxent_objective}.

\subsection{Reinforcement Learning via Variational Inference}
\begin{figure}[t]
    \centering
  	\subfigure[]{
    \label{fig:pgm_no_policy}
    \includegraphics[width=0.15\textwidth]{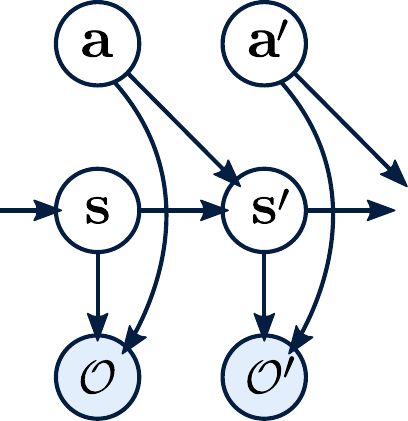}}
    \hfill
  	\subfigure[]{
    \label{fig:pgm_with_policy}
    \includegraphics[width=0.15\textwidth]{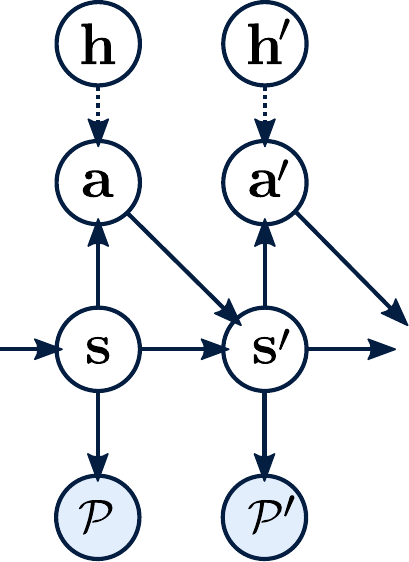}}
    \hfill
  	\subfigure[]{
    \label{fig:pgm_with_latent}
    \includegraphics[width=0.15\textwidth]{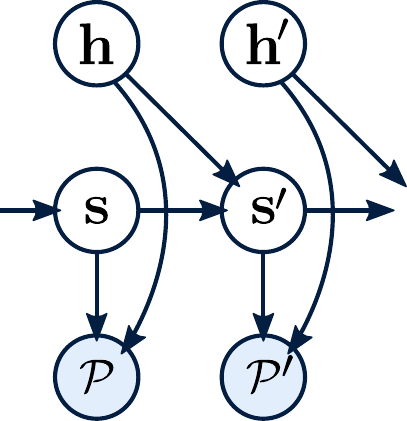}}
    \caption{\small (a) The optimal control problem can be cast as an inference problem by considering a graphical model that consist of transition probabilities, action priors, and optimality variables. We can infer the optimal actions by conditioning on the optimality variables. (b) We can then train a latent variable policy to approximate the optimal actions, augment the graphical model with the policy's action distribution, and condition on a new set of optimality variables $\mathcal{P}_{0:T}$. Dashed line denotes a deterministic (invertible) edge. (c) By marginalizing out the actions $\at$, we are left with a new model that is structurally identical to the one in (a), where $\latentt$ has taken the role of the original actions.}
 \end{figure}

The optimal action distribution inferred from \autoref{eq:optimal_trajectory_distribution} cannot be directly used as a policy for two reasons. First, it would lead to an overly optimistic policy that assumes that the stochastic state transitions can also be modified to prefer optimal behavior, even though in practice, the agent has no control over the dynamics. Second, in continuous domains, the optimal policy is intractable and has to be approximated, for example by using a Gaussian distribution. We can correct both issues by using structured variational inference, where we approximate the posterior with a probabilistic model that constrains the dynamics and the policy. We constrain the dynamics in this distribution to be equal to the true dynamics, which we do not need to actually know in practice but can simply sample in model-free fashion, and constrain the policy to some parameterized distribution. This defines the variational distribution $q(\tau)$ as
\begin{align}
q(\tau) = p(\sz)\prod_{t=0}^{T} \policy(\at|\st) p(\stp|\st, \at),
\end{align}
where $p(\sz)$ and $p(\stp|\st,\at)$ are the true initial state distribution and dynamics, and $\policy(\at|\st)$ is the parameterized policy that we wish to learn.
We can fit this distribution by maximizing the evidence lower bound (ELBO):
\begin{align}
\log p(\optim_{0:T}) %
 &\geq -\kl{q(\tau)}{p(\optim_{0:T}, \tau)}.
\end{align}
Since the dynamics and initial state distributions in $q$ and $p$ match, it's straightforward to check that the divergence term simplifies to
\begin{align}
\label{eq:constrained_max_ent}
\resizebox{\columnwidth}{!}{$
J(\policy) = \E{\tau \sim \rho_\policy(\tau)}{\displaystyle\sum_{t=0}^{T}\reward(\st, \at)\!-\!\kl{\policy(\voidarg|\st)}{p(\voidarg)}}$,}
\end{align}
which, if we choose a uniform action prior, is exactly the maximum entropy objective in \autoref{eq:maxent_objective} up to a constant, and it can be optimized with any off-the-shelf entropy maximizing reinforcement learning algorithms (\eg,~\citep{nachum2017bridging, schulman2017equivalence, haarnoja2017reinforcement, haarnoja2018soft}). Although the variational inference framework is not the only way to motivate the maximum entropy RL objective, it provides a convenient starting point for our method, which will augment the graphical model in \autoref{fig:pgm_no_policy} with latent variables that can then be used to produce a hierarchy of policies, as discussed in the following section.

\section{Learning Latent Space Policies}
\label{sec:our_method}

In this section, we discuss how the probabilistic view of RL can be employed in the construction of hierarchical policies, by augmenting the graphical model in \autoref{fig:pgm_no_policy} with latent variables. We will also propose a particular way to parameterize the distribution over actions conditioned on these latent variables that is based on bijective transformations, which will provide us with a model amenable to stable and tractable training and the ability for higher levels in the hierarchy to fully invert the behavior of the lower layers, as we will discuss in \autoref{sec:real_nvp}. We will derive the method for two-layer hierarchies to simplify notation, but it can be easily generalized to arbitrarily deep hierarchies. In this work, we consider the bottom-up approach, where we first train a low-level policy, and then use it to provide a higher-level action space for a higher level policy that ideally can now solve an easier problem. 

\subsection{Latent Variable Policies for Hierarchical RL}
\label{sec:hierarchy_basic}

We start constructing a hierarchy by defining a stochastic base policy as a latent variable model. In other words, we require the base policy to consist of two factors: a conditional action distribution $\policy(\at|\st, \latentt)$, where $\latentt$ is a latent random variable, and a prior $p(\latentt)$. Actions can be sampled from this policy by first sampling $\latentt$ from the prior and then sampling an action conditioned on $\latentt$. Adding the latent variables $\latentt$ results in a new graphical model, which can now be conditioned on some new optimality variables $\mathcal{P}_t$ that can represent either the same task, or a different higher-level task, as shown in \autoref{fig:pgm_with_policy}. In this new graphical model, the base policy is integrated into the transition structure of the MDP, which now exposes a new, higher-level set of actions $\latentt$. Insofar as the base policy succeeds in solving the task, partially or completely, learning a related task with $\latentt$ as the action should be substantially easier.

This ``policy-augmented'' graphical model has a semantically identical interpretation as the original graphical model in \autoref{fig:pgm_no_policy}, where the combination of the transition model and the action conditional serves as a new (and likely easier) dynamical system.
We can derive the dynamics model for the combined system by marginalizing out the actions: $p(\stp|\st,\latentt)=\int_\aspace p(\stp|\st,\at)\policy(\at|\st,\latentt)d\at$, where $\latentt$ is a new, higher-level action and $p(\latentt)$ is its prior, as illustrated in \autoref{fig:pgm_with_latent}. In other words, the base policy shapes the underlying dynamics of the system, ideally making it more easily controllable by a higher level policy. We can now learn a higher level policy for the latents by conditioning on new optimality variables. We can repeat this process multiple times by integrating each new policy level into the dynamics and learning a new higher-level policy on top of the previous policy's latent space, thus constructing an arbitrarily deep hierarchical policy representation. We will refer to these policy layers as sub-policies in the following sections. Similar layerwise training has been studied in the context of generative modeling with deep believe networks and shown to improve optimization of deep architectures~\cite{hinton2006reducing}.

\subsection{Practical Training of Latent Variable Policies}
\label{sec:real_nvp}

An essential choice in our method is the representation of sub-policies which, ideally, should be characterized by three properties. First, each sub-policy should be tractable. This is required, since we need to maximize the log-likelihood of good actions, which requires marginalizing out all latent variables. Second, the sub-policies should be expressive, so that they don't suppress the information flow from the higher levels to the lower levels. For example, a mixture of Gaussians as a sub-policy can only provide a limited number of behaviors for the higher levels, corresponding to the mixture elements, potentially crippling the ability of the higher layers to solve the task. Third, the conditional factor of each sub-policy should be deterministic, since the higher levels view it as a part of the environment, and additional implicit noise in the system can degrade their performance. Our approach to model the conditionals is based on bijective transformations of the latent variables into actions, which provides all of the aforementioned properties: the sub-policies are tractable, since marginalization reduces to single-point evaluation; they are expressive if represented via neural networks; and they are deterministic due to the bijective transformation. Specifically, we borrow from the recent advances in unsupervised learning based on real-valued non-volume preserving (real NVP) neural network transformations~\citep{dinh2016density}. Our network differs from the original real NVP architecture in that we also condition the transformations on the current state or observation. Note that, even though the transformation from the latent to the action is bijective, it can depend on the observation in arbitrarily complex non-invertible ways, as we discuss in \autoref{sec:architecture}, providing our sub-policies with the requisite expressive power.

We can learn the parameters of these networks by utilizing the change of variables formula as discussed by~\citet{dinh2016density}, and summarized here for completeness. Let \mbox{$\at = f(\latentt;\st)$} be a bijective transformation, and let \mbox{$\latentt\in\reals^{|\aspace|}$} be a random variable and $p(\latentt)$ its prior density. It is possible to express the density of $\at$ in terms of the prior and the Jacobian of the transformation by employing the change of variable formula as 
\begin{align}
\policy(\at|\st) = p(\latentt)\left|\det \left(\frac{df(\latentt;\st)}{d\latentt}\right)\right|\inv.
\end{align}
\citet{dinh2016density} propose a particular kind of bijective transformation, which has a triangular Jacobian, simplifying the computation of the determinant to a product of its diagonal elements. The exact structure of these transformations is outside of the scope of this work, and we refer the reader to \citep{dinh2016density} for a more detailed description. We can easily chain these transformations to form multi-level policies, and we can train them end-to-end as a single policy or layerwise as a hierarchical policy. As a consequence, the policy representation is agnostic to whether or not it was trained as a single expressive latent-variable policy or as a hierarchical policy consisting of several sub-policies, allowing us to choose the training method that best suits the problem at hand without the need to redesign the policy topology each time from scratch. Next, we will discuss the different hierarchical training strategies in more detail.

\subsection{Reward Functions for Policy Hierarchies}

The simplest way to construct a hierarchy out of latent variable policies is to follow the procedure described in~\autoref{sec:hierarchy_basic}, where we train each layer in turn, then freeze its weights, and train a new layer that uses the lower layer's latent variables as an action space. In this procedure, each layer is trained on the same maximum entropy objective, and each layer simplifies the task for the layer above it. As we will show in~\autoref{sec:experiments}, this procedure can provide substantial benefit on challenging and high-dimensional benchmark tasks.

However, for tasks that are more challenging, we can also naturally incorporate weak prior information into the training process by using underdefined heuristic reward functions for training the lower layers. In this approach, lower layers of the hierarchy are trained on reward functions that include shaping terms that elicit more desirable behaviors. For example, if we wish to learn a complex navigation task for a walking robot, the lower-layer objective might provide a reward for moving in any direction, while the higher layer is trained only on the primary objective of the task. Because our method uses bijective transformations, the higher layer can always undo any behavior in the lower layer if it is detrimental to task success, while lower layer objectives that correlate with task success can greatly simplify learning. Furthermore, since each layer still aims to maximize entropy, even a weak objective, such as a reward for motion in any direction (e.g., the norm of the velocity), will produce motion in many different directions that is controllable by the lower layer's latent variables. In our experiments, we will demonstrate how this approach can be used to solve a goal navigation task for a simulated, quadrupedal robot.

\subsection{Algorithm Summary}

We summarize the proposed algorithm in~\autoref{alg:algorithm}. The algorithm begins by operating on the low-level actions in the environment according to the unknown system dynamics $p(\stp|\st,\at)$, where we use $\latentt^{(0)} = \at$ to denote the lowest-layer actions. The algorithm is also provided with an ordered set of $K$ reward functions $\mathcal{R}_i$, which can all represent the same task or different tasks, depending on the skill we want to learn: skills that naturally divide into primitive skills can benefit from specifying a different low-level objective, but for simpler tasks, such as locomotion, which do not naturally divide into primitives, we can train each sub-policy to optimize the same objective. In both cases, the last reward function $\mathcal{R}_{K-1}$ should correspond to the actual task we want to solve. The algorithm then chooses each $\mathcal{R}_i$ sequentially and learns a maximum entropy policy $\policy_i$, represented by an invertible transformation $f_i: \sspace \times \aspace \rightarrow \aspace$ and a prior $p(\latentt^{(i)})$, to optimize the corresponding variational inference objective in \autoref{eq:constrained_max_ent}. Our proposed implementation uses soft actor-critic~\citep{haarnoja2018soft} to optimize the policy due to its robustness and  good sample-efficiency, although other entropy maximizing RL algorithms can also be used. After each iteration, we embed the newly learned transformation $f_i$ into the environment, which produces a new system dynamics $p(\stp|\st,\latentt^{(i+1)})$ that can be used by the next layer. As before, we do not need an analytic form of these dynamics, only the ability to sample from them, which allows our algorithm to operate in the fully model-free setting.

\renewcommand{\algorithmicrequire}{\textbf{Input:}}
\renewcommand{\algorithmicensure}{\textbf{Output:}}
\begin{algorithm}[tbh]
\caption{Latent Space Policy Learning}
\label{alg:algorithm}
\begin{algorithmic}
\REQUIRE True environment $p(\stp|, \st, \latentt^{(0)})$, where $\latentt^{(0)}$ 
\STATE corresponds to the physical actions $\at$.
\REQUIRE Reward specifications $\{\mathcal{R}_0, \mathcal{R}_1,..., \mathcal{R}_{K-1}\}$.
\FOR{$i = 0$ to $K-1$}
	\STATE Initialize the weights of layer $f_i$.
	\STATE Learn the weights of $f_i$ so that $\latentt^{(i)} = f_i(\latentt^{(i+1)}; \st)$, 
	\STATE \hspace{3mm} where $\latentt^{(i+1)} \sim p(\latentt^{(i+1)})$, optimizes $\mathcal{R}_i$
	\STATE \hspace{3mm} on $p(\stp|\st, \latentt^{(i)})$.
	\STATE Embed the new layer $f_i$ into the environment:
	\STATE \hspace{3mm}  $p(\stp|\st, \latentt^{(i+1)}) \leftarrow p(\stp|\st, f_i(\latentt^{(i+1)}; \st))$.
\ENDFOR
\ENSURE A hierarchical policy $f = f_0 \circ f_1 \circ ... \circ f_{K-1}$.
\end{algorithmic}
\end{algorithm}

\section{Experiments}
\label{sec:experiments}

Our experiments were conducted on several continuous control benchmark tasks from the OpenAI Gym benchmark suite~\cite{brockman2016openai}. The aim of our experiments was to answer the following questions: (1) How well does our latent space policy learning method compare to prior reinforcement learning algorithms? (2) Can we attain improved performance from adding additional latent variable policy layers to a hierarchy, especially on challenging, high-dimensional tasks? (3) Can we solve more complex tasks by providing simple heuristic shaping to lower layers of the hierarchy, while the higher layers optimize the original task reward? Videos of our experiments are available online\footnote{\href{https://sites.google.com/view/latent-space-deep-rl}{https://sites.google.com/view/latent-space-deep-rl}}.

\subsection{Policy Architecture}
\label{sec:architecture}

\begin{figure}[t]
\center
    \includegraphics[width=0.60\columnwidth]{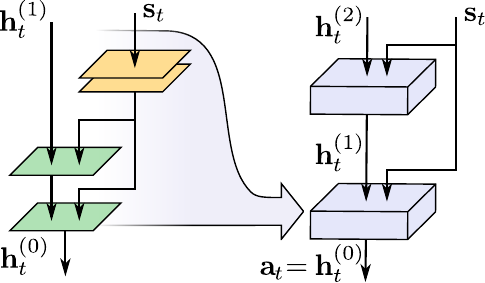}
    \caption{\small Our hierarchical policy consist of two levels (right diagram) that take in the observation and a latent vector from the previous level and outputs a latent vector to the next level. Diagram on the left shows the internal structure of each of the policy levels. The latent vector that is passed from the higher level is fed through two invertible coupling layers (green)~\citep{dinh2016density}, which we condition on an observation embedding (yellow). Note that the path from the observation to the output does not need to be bijective, and therefore the observation embeddings can be represented with an arbitrary neural network, which in our case consists of two fully connected layers.}
    \label{fig:network}
\end{figure}

In our experiments, we used both single-level policies and hierarchical policies composed of two sub-policies as shown on the right in \autoref{fig:network}. Each sub-policy has an identical structure, as depicted on the left in \autoref{fig:network}. A sub-policy is constructed from two \emph{coupling layers} that are connected using the alternating pattern described in~\citep{dinh2016density}. Our implementation differs from \citet{dinh2016density} only in that we condition the coupling layers on the observations, via an embedding performed by a two-layer fully-connected network. In practice, we concatenate the embedding vector with the latent input to each coupling layer. Note that our method requires only the path from the input latent to the output to be invertible, and the output can depend on the observation in arbitrarily complex ways. We have released our code for reproducibility.\footnote{\href{https://github.com/haarnoja/sac}{https://github.com/haarnoja/sac}}

\begin{figure}[t]
    \vspace{-2mm}
    \centering
    \hspace{-3mm}
	\subfigure[Swimmer (rllab)]{
    \includegraphics[width=0.24\textwidth, trim={1mm 0 5mm 7.5mm}, clip]{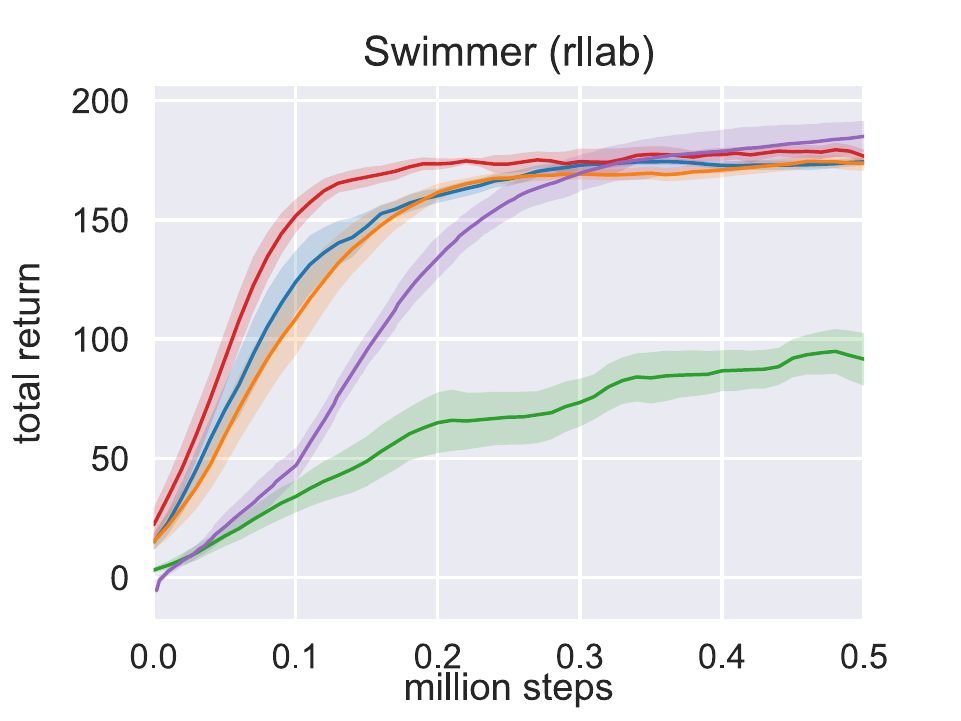}}
    \hspace{-3mm}
    \subfigure[Hopper-v1]{
    \includegraphics[width=0.24\textwidth, trim={1mm 0 5mm 7.5mm}, clip]{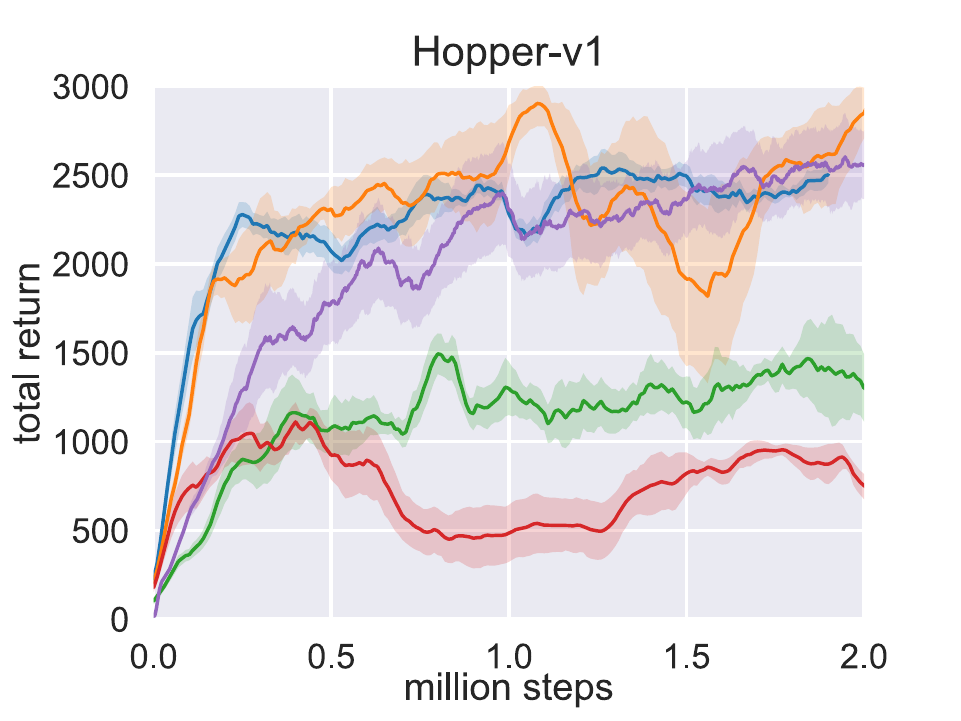}}\\

    \hspace{-3mm}
  	\subfigure[Walker2d-v1]{
    \includegraphics[width=0.24\textwidth, trim={1mm 0 5mm 7.5mm}, clip]{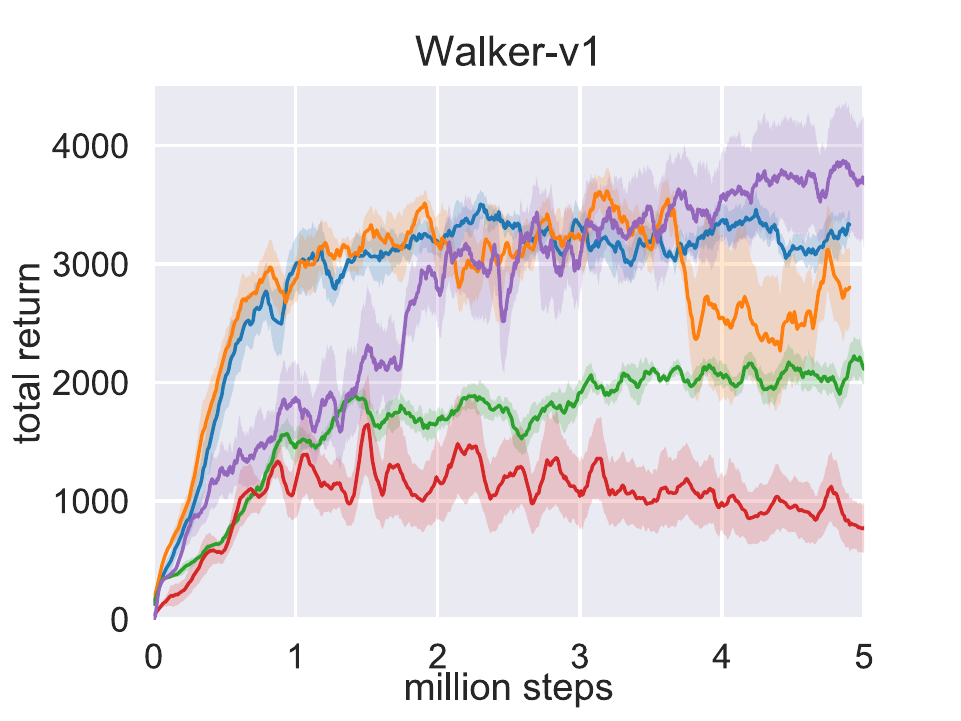}}
    \hspace{-3mm}
    \subfigure[HalfCheetah-v1]{
    \includegraphics[width=0.24\textwidth, trim={1mm 0 5mm 7.5mm}, clip]{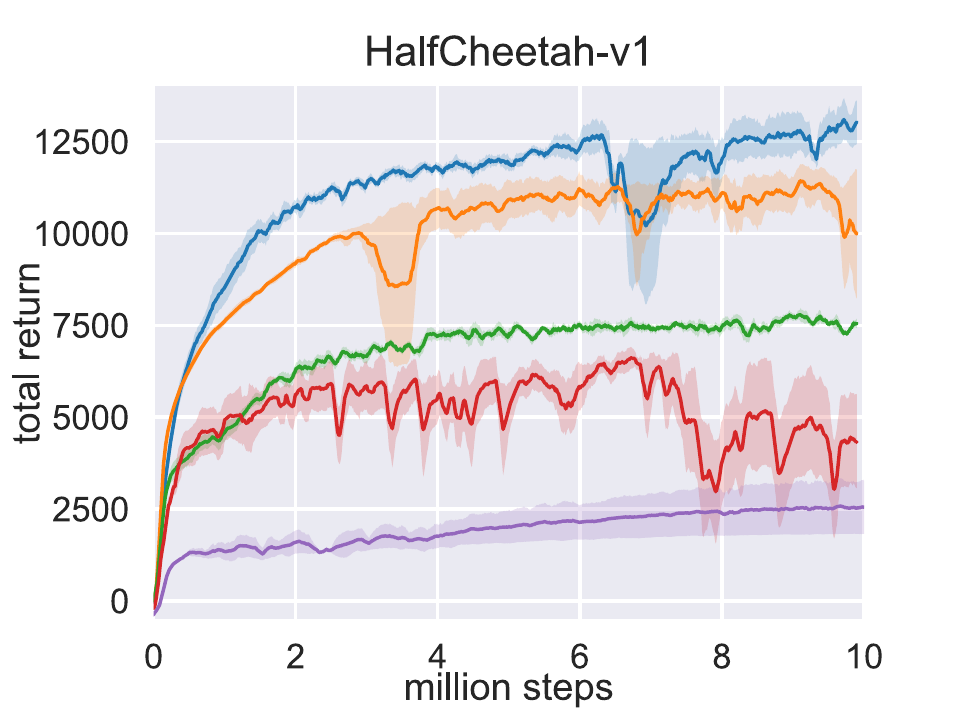}}\\

    \hspace{-3mm}
    \subfigure[Ant (rllab)]{
    \includegraphics[width=0.24\textwidth, trim={1mm 0 5mm 7.5mm}, clip]{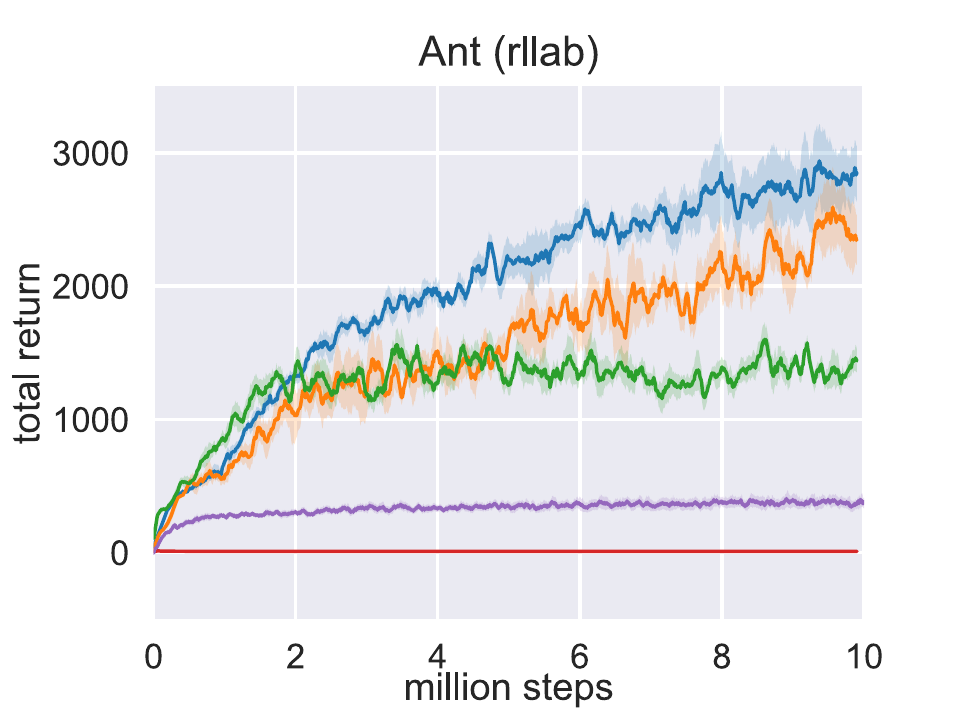}}
    \hspace{-3mm}
    \subfigure[Humanoid (rllab)]{
    \includegraphics[width=0.24\textwidth, trim={1mm 0 5mm 7.5mm}, clip]{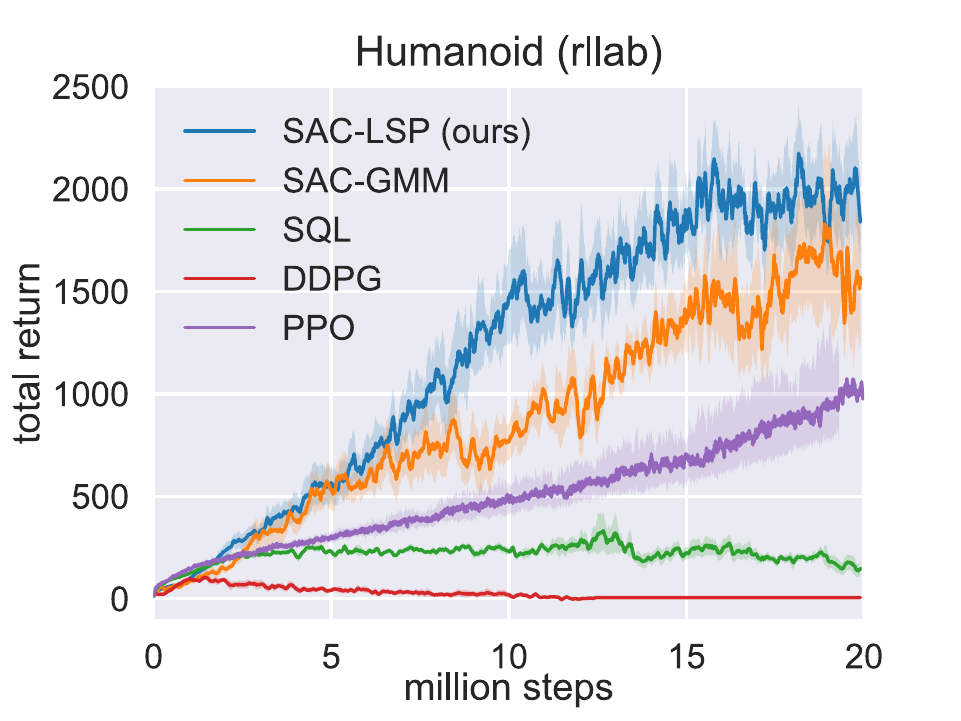}}\\

    \caption{\small Training curves for continuous control benchmarks. Thick lines correspond to mean performance, and shaded regions show standard deviations of five random seeds. Our method (SAC-LSP) attains state-of-the-art performance across all tasks.}
    \vspace{-4mm}
	\label{fig:benchmarks}
 \end{figure}

\subsection{Benchmark Tasks with Single-Level Policies}
\vspace{-1mm}

We compare our method (SAC-LSP)\textbf{} to two commonly used RL algorithms: proximal policy optimization (PPO)~\citep{schulman2017proximal}, a commonly used policy gradient algorithm, and deep deterministic policy gradient (DDPG)~\citep{lillicrap2015continuous}, which is a sample efficient off-policy method.
We also include two recent algorithm that learn maximum entropy policies: soft Q-learning (SQL)~\citep{haarnoja2017reinforcement}, which also learns a sampling network as part of the model, represented by an implicit density model, and soft actor-critic~\citep{haarnoja2018soft}, which uses a Gaussian mixture model policy (SAC-GMM). Note that the benchmark tasks compare the total expected return, but the entropy maximizing algorithms optimize a slightly different objective, so this comparison slightly favors PPO and DDPG, which optimize the benchmark objective directly. Another difference between the two classes of algorithms is that the maximum entropy policies are stochastic at test time, while DDPG is deterministic and PPO typically converges to nearly deterministic policies. For SAC-GMM, we execute an approximate maximum a posteriori action by choosing the mean of the mixture component that has the highest Q-value at test time, but for SQL and our method, which both can represent an arbitrarily complex posterior distribution, we simply sample from the stochastic policy.

Our results on the benchmark tasks show that our policy representation generally performs on par or better than all of the tested prior methods, both in terms of efficiency and final return (\autoref{fig:benchmarks}), especially on the more challenging and high-dimensional tasks, such as Humanoid. These results indicate that our policy representation can accelerate learning, particularly in challenging environments with high-dimensional actions and observations.

\begin{figure}[t]
    \vspace{-2mm}
    \centering
    \hfill
  	\subfigure[Ant (rllab)]{
    \label{fig:benchmark_stack_ant}
    \includegraphics[width=0.23\textwidth, trim={1mm, 0, 5mm, 7.5mm}, clip]{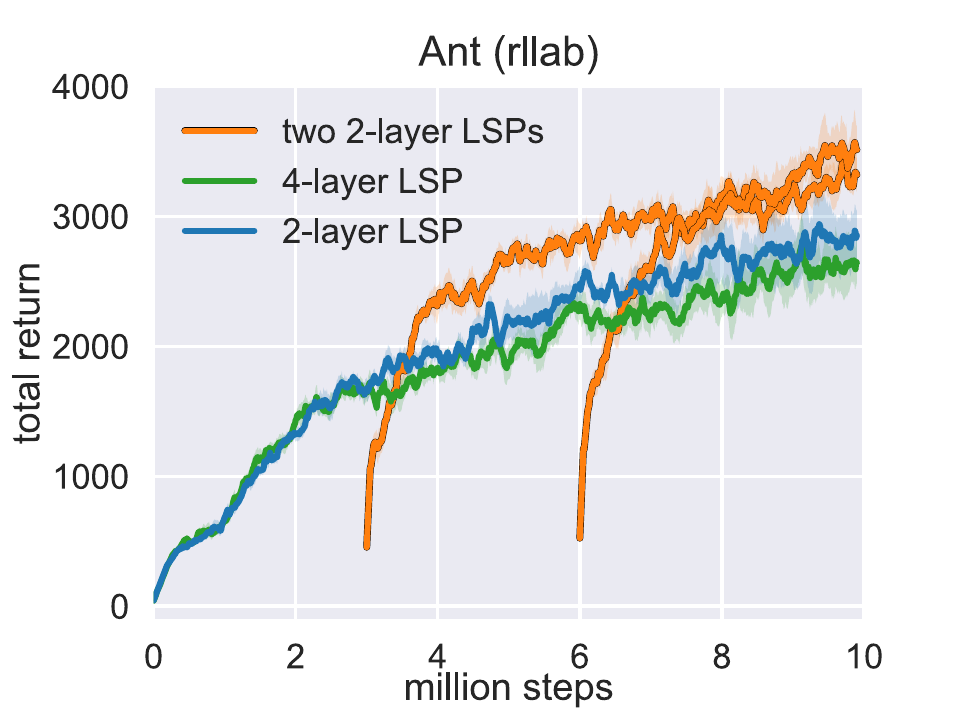}}
    \hfill
  	\subfigure[Humanoid (rllab)]{
    \label{fig:benchmark_stack_humanoid}
    \includegraphics[width=0.23\textwidth, trim={1mm, 0, 5mm, 7.5mm}, clip]{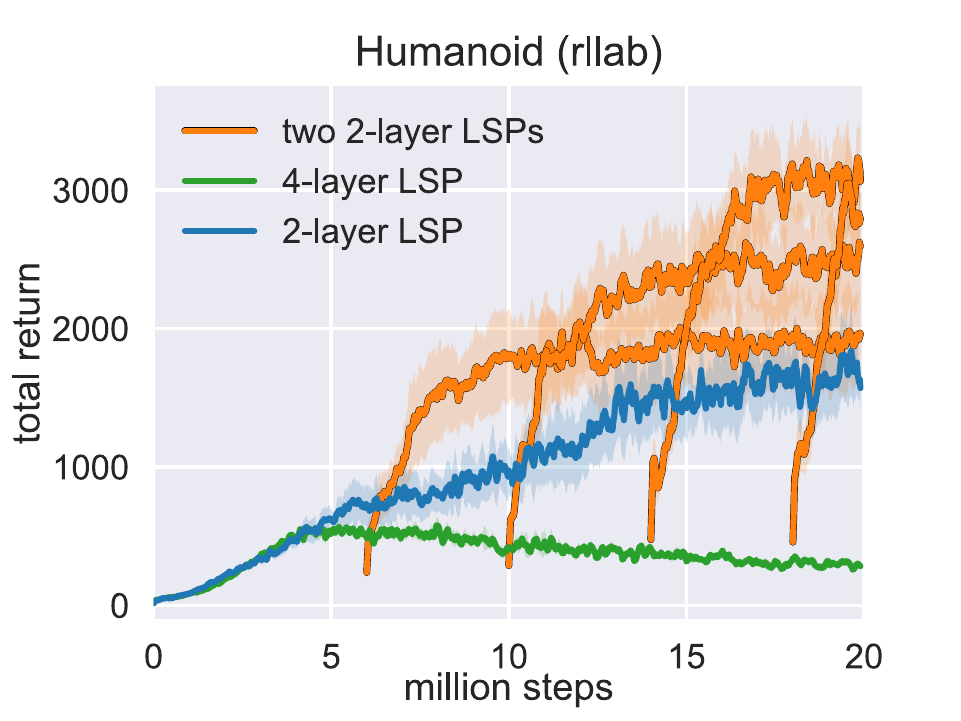}}
    \hfill
    \vspace{-1mm}
    \caption{\small (a, b) We trained two-level policies for the most challenging benchmark tasks, Ant and Humanoid, by first training a single level policy (blue) and then freezing it and adding a second policy level. We repeated this procedure by starting training of the top-level policy at multiple points in time (orange). We also trained a single policy with four invertible layers end-to-end (green) for comparison. In each case, stagewise training of two levels yields the best performance.}
    \vspace{-5mm}
 \end{figure}

\vspace{-1mm}
\subsection{Multi-Level Policies}
\vspace{-1mm}

In this section, we evaluate the performance of our approach when we compose multiple latent variable policies into a hierarchy. In the first experiment, we train a single-level base policy on the most challenging standard benchmarks, Ant and Humanoid. We then freeze the weights of the base policy, and learn another policy level that uses the latent variables of the first policy as its action space. Intuitively, each layer in such a hierarchy attempts to solve the task to the best of its ability, providing an easier problem for the layer above. In \autoref{fig:benchmark_stack_ant} and \autoref{fig:benchmark_stack_humanoid}, we show the training curves for Ant and Humanoid, where the blue curve corresponds to the base policy and orange curves show the performance after we freeze the base policy weights and optimize a second-level policy. The different orange curves correspond to the addition of the second layer after a different numbers of training steps. In each case, the two-level policy can outperform a single level policy by a large margin. The performance boost is more prominent if we train the base policy longer. Note that the base policy corresponds to a single-level policy in \autoref{fig:benchmarks}, and already learns more efficiently than prior methods. We also compare to a single, more expressive policy (green) that consists of four invertible layers, which has a similar number of parameters to a stacked two-level policy, but trained end-to-end as oppose to stagewise. This single four-layer policy performs comparably (Ant) or worse (Humanoid) than a single, two-layer policy (blue), indicating that the benefit of the two-level hierarchy is not just in the increased expressivity of the policy, but that the stagewise training procedure plays an important role in improving performance.

\begin{figure}[t]
    \vspace{-2mm}
     \centering
     \hfill
     \subfigure[Ant maze]{
     \label{fig:ant_maze_env}
     \includegraphics[width=0.23\textwidth, trim={4cm, -25.5mm, 4cm, 22mm}, clip]{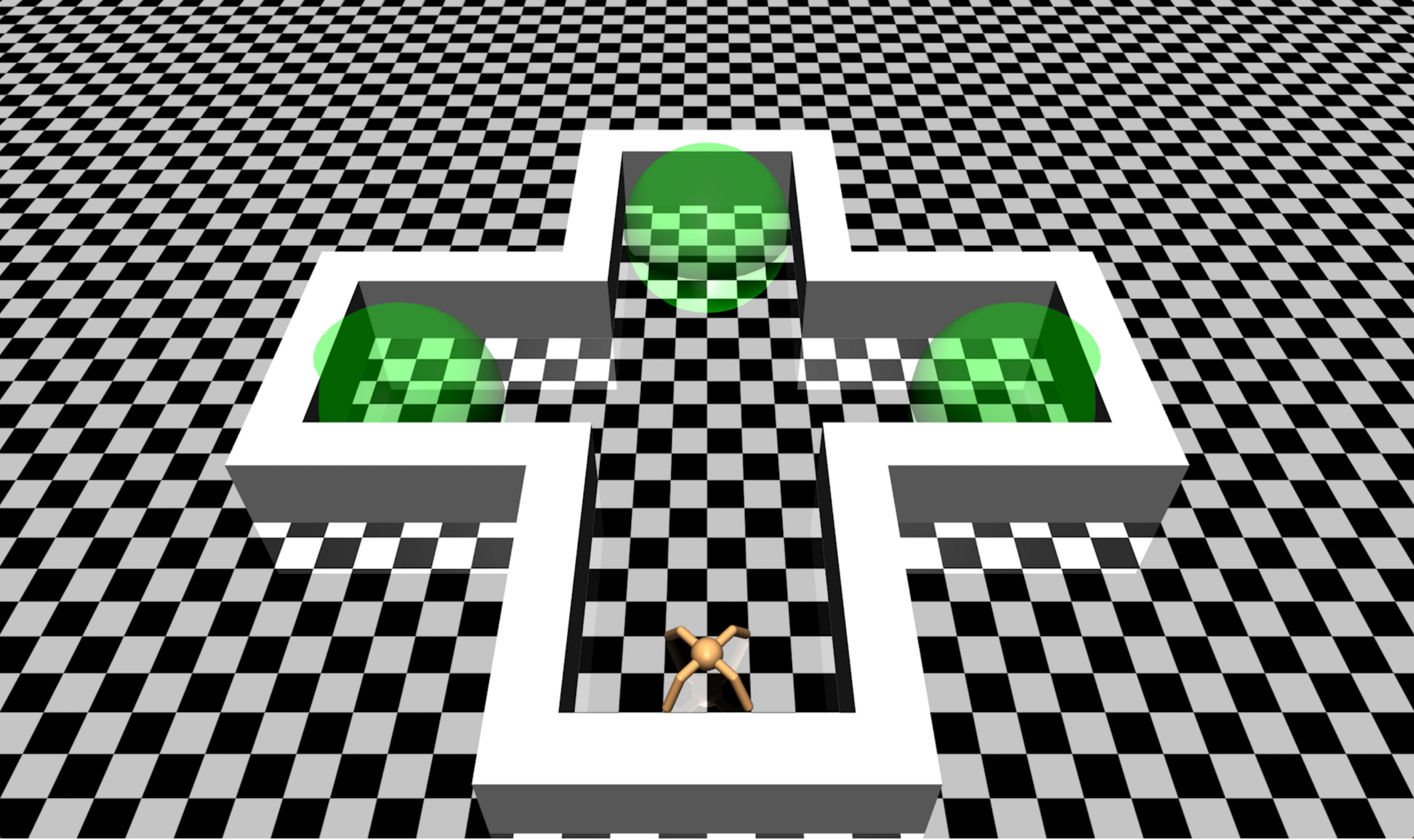}}
     \hfill
     \subfigure[Ant maze results]{
     \label{fig:ant_hierarchy}
     \includegraphics[width=0.23\textwidth, trim={2mm, 0, 0mm, 0mm}, clip]{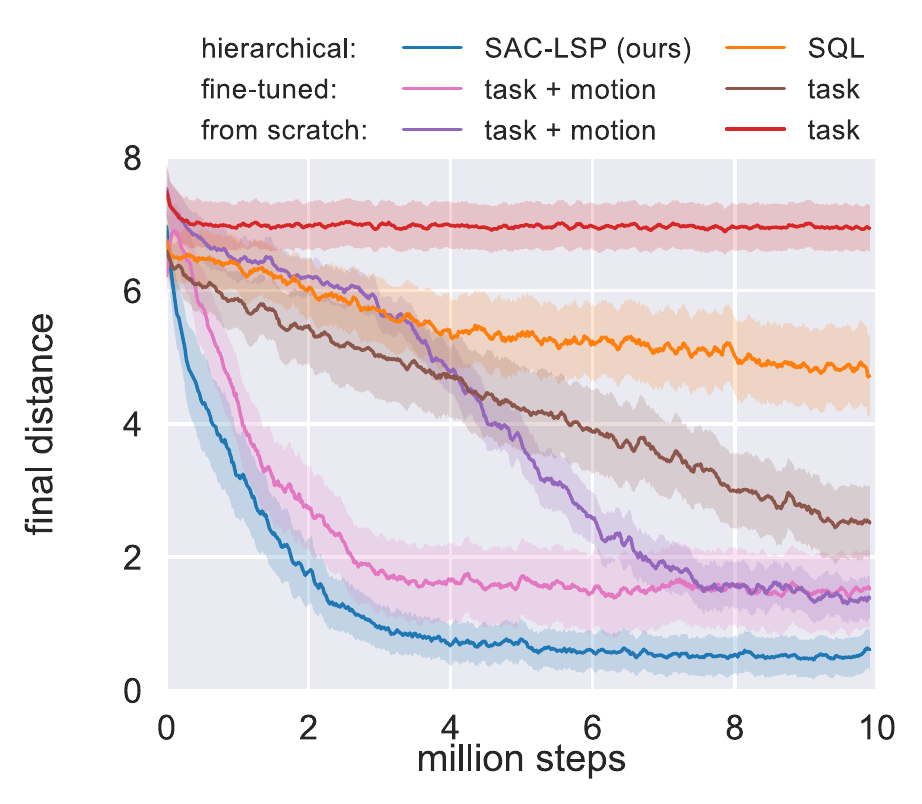}}
     \hfill
     \caption{\small (a) We trained Ant to navigate through a simple maze to three different goal location shown in green. (b) We first trained a low-level policy with a motion reward in a pretraining environment not including the walls, then fixed the policy and trained another policy level with a target reward (blue). We compare our method to learning a single policy from scratch or fine-tuning the pretrained policy using only the task reward or a combination of task and motion rewards. We also applied our method to soft Q-learning.}
    \vspace{-4mm}
 \end{figure}

In our second experiment, we study how our method can be used to build hierarchical policies for complex tasks that require compound skills. The particular task we consider requires Ant to navigate through a simple maze (\autoref{fig:ant_maze_env}) with a sparse, binary reward for reaching the goals (shown in green). To solve this task, we can construct a hierarchy, where the lower layer is trained to acquire a general locomotion skill simply by providing a reward for maximizing velocity, regardless of direction. This pretraining phase can be conducted in a simpler, open environment where the agent can move freely. This provides a small amount of domain knowledge, but this type of domain knowledge is much easier to specify than true intermediate goals or modulation~\citep{heess2016learning,kulkarni2016hierarchical}, and the entropy maximization term in the objective automatically causes the low-level policy to learn a range of locomotion skills for various directions. The higher-level policy is then provided the standard task reward. Because the lower-level policy is invertible, the higher-level policy can still solve the task however it needs to, potentially even by fully undoing the behavior of the lower-level policy. However, the results suggest that the lower-level policy does indeed substantially simplify the problem, resulting in both rapid learning and good final performance.

In \autoref{fig:ant_hierarchy}, we compare our approach (blue) to four single-level baselines that either learn a policy from scratch or fine-tune a pretrained policy. The pretraining phase (4 million steps) is not included in the learning curves, since the same base policies were reused across multiple tasks, corresponding to the three difference goal locations shown in~\autoref{fig:ant_maze_env}. With task reward only, training a policy from scratch (red) failed to solve the task due to lack of structured exploration, whereas fine-tuning a pretrained policy (brown) that already knows how to move around and could occasionally find the way to the goal was able to slowly learn this task. We also tried improving exploration by augmenting the objective with a motion reward, provided as a shaping term for the entire policy. In this case, a policy trained from scratch (purple) learned slowly, as it first needed to acquire a locomotion skill, but was able to eventually solve the task, while fine-tuning a pretrained policy resulted in much faster learning (pink). However, in both cases, adding motion as a shaping term prevents the policy from converging on an optimal solution, since the shaping alters the task. This manifests as residual error at convergence. On the other hand, our method (blue) can make use of the pretrained locomotion skills while optimizing for the task reward directly, resulting in faster learning and better final performance. Note that our method converges to a solution where the final distance to the goal is more than four times smaller than for the next best method, which finetunes with a shaped reward. We also applied our method to soft Q-learning (yellow), which also trains latent space policies, but we found it to learn substantially slower than SAC-LSP.

\vspace{-1mm}
\section{Discussion and Future Work}
\vspace{-1mm}

We presented a method for training policies with latent variables. Our reinforcement learning algorithm not only compares favorably to state-of-the-art algorithms on standard benchmark tasks, but also provides an appealing avenue for constructing hierarchical policies: higher levels in the hierarchy can directly make use of the latent space of the lower levels as their action space, which allows us to train the entire hierarchy in a layerwise fashion. This approach to hierarchical reinforcement learning has a number of conceptual and practical benefits. First, each layer in the hierarchy can be trained with exactly the same algorithm. Second, by using an invertible mapping from latent variables to actions, each layer becomes invertible, which means that the higher layer can always perfectly invert any behavior of the lower layer. This makes it possible to train lower layers on heuristic shaping rewards, while higher layers can still optimize task-specific rewards with good asymptotic performance. Our method has a natural interpretation as an iterative procedure for constructing graphical models that gradually simplify the task dynamics. 
Our approach can be extended to enable different layers to operate at different temporal scales, which would provide for extensions into domains with temporally delayed rewards and multi-stage tasks.

\section*{Acknowledgments}
We thank Aurick Zhou for producing some of the baseline results. This work was supported by Siemens and Berkeley DeepDrive.
\FloatBarrier

\bibliography{refs}

\begin{thebibliography}{39}
\providecommand{\natexlab}[1]{#1}
\providecommand{\url}[1]{\texttt{#1}}
\expandafter\ifx\csname urlstyle\endcsname\relax
  \providecommand{\doi}[1]{doi: #1}\else
  \providecommand{\doi}{doi: \begingroup \urlstyle{rm}\Url}\fi

\bibitem[Bacon et~al.(2017)Bacon, Harb, and Precup]{bacon2017option}
Bacon, P.-L., Harb, J., and Precup, D.
\newblock The option-critic architecture.
\newblock In \emph{AAAI Conference on Artificial Intelligence (AAAI)}, pp.\
  1726--1734, 2017.

\bibitem[Brockman et~al.(2016)Brockman, Cheung, Pettersson, Schneider,
  Schulman, Tang, and Zaremba]{brockman2016openai}
Brockman, G., Cheung, V., Pettersson, L., Schneider, J., Schulman, J., Tang,
  J., and Zaremba, W.
\newblock Open{AI} {G}ym.
\newblock \emph{arXiv preprint arXiv:1606.01540}, 2016.

\bibitem[Daniel et~al.(2012)Daniel, Neumann, and
  Peters]{daniel2012hierarchical}
Daniel, C., Neumann, G., and Peters, J.
\newblock Hierarchical relative entropy policy search.
\newblock In \emph{Artificial Intelligence and Statistics}, pp.\  273--281,
  2012.

\bibitem[Dinh et~al.(2016)Dinh, Sohl-Dickstein, and Bengio]{dinh2016density}
Dinh, L., Sohl-Dickstein, J., and Bengio, S.
\newblock Density estimation using {R}eal {NVP}.
\newblock \emph{arXiv preprint arXiv:1605.08803}, 2016.

\bibitem[Eysenbach et~al.(2018)Eysenbach, Gupta, Ibarz, and
  Levine]{eysenbach2018diversity}
Eysenbach, B., Gupta, A., Ibarz, J., and Levine, S.
\newblock Diversity is all you need: Learning skills without a reward function.
\newblock \emph{arXiv preprint arXiv:1802.06070}, 2018.

\bibitem[Florensa et~al.(2017)Florensa, Duan, and P.]{florensa2017stochastic}
Florensa, C., Duan, Y., and P., A.
\newblock Stochastic neural networks for hierarchical reinforcement learning.
\newblock In \emph{International Conference on Learning Representations
  (ICLR)}, 2017.

\bibitem[Fox et~al.(2016)Fox, Pakman, and Tishby]{fox2015taming}
Fox, R., Pakman, A., and Tishby, N.
\newblock Taming the noise in reinforcement learning via soft updates.
\newblock In \emph{Conference on Uncertainty in Artificial Intelligence}, 2016.

\bibitem[Haarnoja et~al.(2017)Haarnoja, Tang, Abbeel, and
  Levine]{haarnoja2017reinforcement}
Haarnoja, T., Tang, H., Abbeel, P., and Levine, S.
\newblock Reinforcement learning with deep energy-based policies.
\newblock In \emph{International Conference on Machine Learning (ICML)}, pp.\
  1352--1361, 2017.

\bibitem[Haarnoja et~al.(2018)Haarnoja, Zhou, Abbeel, and
  Levine]{haarnoja2018soft}
Haarnoja, T., Zhou, A., Abbeel, P., and Levine, S.
\newblock Soft actor-critic: Off-policy maximum entropy deep reinforcement
  learning with a stochastic actor.
\newblock \emph{arXiv preprint arXiv:1801.01290}, 2018.

\bibitem[Hausman et~al.(2018)Hausman, Springenberg, Wang, Heess, and
  Riedmiller]{hausman2018learning}
Hausman, K., Springenberg, J.~T., Wang, Z., Heess, N., and Riedmiller, M.
\newblock Learning an embedding space for transferable robot skills.
\newblock In \emph{Conference on Learning Representations (ICLR)}, 2018.

\bibitem[Heess et~al.(2016)Heess, Wayne, Tassa, Lillicrap, Riedmiller, and
  Silver]{heess2016learning}
Heess, N., Wayne, G., Tassa, Y., Lillicrap, T., Riedmiller, M., and Silver, D.
\newblock Learning and transfer of modulated locomotor controllers.
\newblock \emph{arXiv preprint arXiv:1610.05182}, 2016.

\bibitem[Hinton \& Salakhutdinov(2006)Hinton and
  Salakhutdinov]{hinton2006reducing}
Hinton, G.~E. and Salakhutdinov, R.~R.
\newblock Reducing the dimensionality of data with neural networks.
\newblock \emph{Science}, 313\penalty0 (5786):\penalty0 504--507, 2006.

\bibitem[Kappen(2005)]{kappen2005path}
Kappen, H.~J.
\newblock Path integrals and symmetry breaking for optimal control theory.
\newblock \emph{Journal of Statistical Mechanics: Theory And Experiment},
  2005\penalty0 (11):\penalty0 P11011, 2005.

\bibitem[Kulkarni et~al.(2016)Kulkarni, Narasimhan, Saeedi, and
  Tenenbaum]{kulkarni2016hierarchical}
Kulkarni, T.~D., Narasimhan, K., Saeedi, A., and Tenenbaum, J.
\newblock Hierarchical deep reinforcement learning: Integrating temporal
  abstraction and intrinsic motivation.
\newblock In \emph{Advances in Neural Information Processing Systems (NIPS)},
  pp.\  3675--3683, 2016.

\bibitem[Kupcsik et~al.(2013)Kupcsik, Deisenroth, Peters, and
  Neumann]{kupcsik2013data}
Kupcsik, A.~G., Deisenroth, M.~P., Peters, J., and Neumann, G.
\newblock Data-efficient generalization of robot skills with contextual policy
  search.
\newblock In \emph{Proceedings of the 27th AAAI Conference on Artificial
  Intelligence (AAAI)}, pp.\  1401--1407, 2013.

\bibitem[Le{C}un et~al.(2015)Le{C}un, Bengio, and Hinton]{lbh-dl-15}
Le{C}un, Y., Bengio, Y., and Hinton, G.
\newblock Deep learning.
\newblock \emph{Nature}, 521:\penalty0 436--444, May 2015.

\bibitem[Levine(2014)]{levine2014motor}
Levine, S.
\newblock \emph{Motor skill learning with local trajectory methods}.
\newblock PhD thesis, Stanford University, 2014.

\bibitem[Levine et~al.(2016)Levine, Finn, Darrell, and Abbeel]{levine2016end}
Levine, S., Finn, C., Darrell, T., and Abbeel, P.
\newblock End-to-end training of deep visuomotor policies.
\newblock \emph{Journal of Machine Learning Research}, 17\penalty0
  (39):\penalty0 1--40, 2016.

\bibitem[Lillicrap et~al.(2015)Lillicrap, Hunt, Pritzel, Heess, Erez, Tassa,
  Silver, and Wierstra]{lillicrap2015continuous}
Lillicrap, T.~P., Hunt, J.~J., Pritzel, A., Heess, N., Erez, T., Tassa, Y.,
  Silver, D., and Wierstra, D.
\newblock Continuous control with deep reinforcement learning.
\newblock \emph{arXiv preprint arXiv:1509.02971}, 2015.

\bibitem[MacAlpine \& Stone(2018)MacAlpine and Stone]{macalpine2018overlapping}
MacAlpine, P. and Stone, P.
\newblock Overlapping layered learning.
\newblock \emph{Artificial Intelligence}, 254:\penalty0 21--43, 2018.

\bibitem[Mnih et~al.(2013)Mnih, Kavukcuoglu, Silver, Graves, Antonoglou,
  Wierstra, and Riedmiller]{mnih2013playing}
Mnih, V., Kavukcuoglu, K., Silver, D., Graves, A., Antonoglou, I., Wierstra,
  D., and Riedmiller, M.
\newblock Playing atari with deep reinforcement learning.
\newblock \emph{arXiv preprint arXiv:1312.5602}, 2013.

\bibitem[Mnih et~al.(2016)Mnih, Badia, Mirza, Graves, Lillicrap, Harley,
  Silver, and Kavukcuoglu]{mnih2016asynchronous}
Mnih, V., Badia, A.~P., Mirza, M., Graves, A., Lillicrap, T.~P., Harley, T.,
  Silver, D., and Kavukcuoglu, K.
\newblock Asynchronous methods for deep reinforcement learning.
\newblock In \emph{International Conference on Machine Learning (ICML)}, 2016.

\bibitem[Nachum et~al.(2017)Nachum, Norouzi, Xu, and
  Schuurmans]{nachum2017bridging}
Nachum, O., Norouzi, M., Xu, K., and Schuurmans, D.
\newblock Bridging the gap between value and policy based reinforcement
  learning.
\newblock In \emph{Advances in Neural Information Processing Systems}, pp.\
  2772--2782, 2017.

\bibitem[Neumann(2011)]{neumann2011variational}
Neumann, G.
\newblock Variational inference for policy search in changing situations.
\newblock In \emph{International Conference on Machine Learning (ICML)}, pp.\
  817--824, 2011.

\bibitem[Peters et~al.(2010)Peters, M{\"u}lling, and Altun]{peters2010relative}
Peters, J., M{\"u}lling, K., and Altun, Y.
\newblock Relative entropy policy search.
\newblock In \emph{AAAI Conference on Artificial Intelligence (AAAI)}, pp.\
  1607--1612, 2010.

\bibitem[Rawlik et~al.(2012)Rawlik, Toussaint, and
  Vijayakumar]{rawlik2012stochastic}
Rawlik, K., Toussaint, M., and Vijayakumar, S.
\newblock On stochastic optimal control and reinforcement learning by
  approximate inference.
\newblock \emph{Robotics: Science and Systems (RSS)}, 2012.

\bibitem[Saxe et~al.(2017)Saxe, Earle, and Rosman]{saxe2017hierarchy}
Saxe, A.~M., Earle, A.~C., and Rosman, B.~S.
\newblock Hierarchy through composition with multitask {LMDP}s.
\newblock \emph{Proceedings of Machine Learning Research}, 2017.

\bibitem[Schaul et~al.(2015)Schaul, Horgan, Gregor, and
  Silver]{schaul2015universal}
Schaul, T., Horgan, D., Gregor, K., and Silver, D.
\newblock Universal value function approximators.
\newblock In \emph{International Conference on Machine Learning (ICML)}, pp.\
  1312--1320, 2015.

\bibitem[Schulman et~al.(2015)Schulman, Levine, Abbeel, Jordan, and
  Moritz]{schulman2015trust}
Schulman, J., Levine, S., Abbeel, P., Jordan, M.~I., and Moritz, P.
\newblock Trust region policy optimization.
\newblock In \emph{International Conference on Machine Learning (ICML)}, pp.\
  1889--1897, 2015.

\bibitem[Schulman et~al.(2017{\natexlab{a}})Schulman, Abbeel, and
  Chen]{schulman2017equivalence}
Schulman, J., Abbeel, P., and Chen, X.
\newblock Equivalence between policy gradients and soft {Q}-learning.
\newblock \emph{arXiv preprint arXiv:1704.06440}, 2017{\natexlab{a}}.

\bibitem[Schulman et~al.(2017{\natexlab{b}})Schulman, Wolski, Dhariwal,
  Radford, and Klimov]{schulman2017proximal}
Schulman, J., Wolski, F., Dhariwal, P., Radford, A., and Klimov, O.
\newblock Proximal policy optimization algorithms.
\newblock \emph{arXiv preprint arXiv:1707.06347}, 2017{\natexlab{b}}.

\bibitem[Silver et~al.(2016)Silver, Huang, Maddison, Guez, Sifre, van~den
  Driessche, Schrittwieser, Antonoglou, Panneershelvam, Lanctot, Dieleman,
  Grewe, Nham, Kalchbrenner, Sutskever, Lillicrap, Leach, Kavukcuoglu, Graepel,
  and Hassabis]{silver2016mastering}
Silver, D., Huang, A., Maddison, C.~J., Guez, A., Sifre, L., van~den Driessche,
  G., Schrittwieser, J., Antonoglou, I., Panneershelvam, V., Lanctot, M.,
  Dieleman, S., Grewe, D., Nham, J., Kalchbrenner, N., Sutskever, I.,
  Lillicrap, T., Leach, M., Kavukcuoglu, K., Graepel, T., and Hassabis, D.
\newblock Mastering the game of go with deep neural networks and tree search.
\newblock \emph{Nature}, 529\penalty0 (7587):\penalty0 484--489, Jan 2016.
\newblock ISSN 0028-0836.
\newblock Article.

\bibitem[Sutton et~al.(1999)Sutton, Precup, and Singh]{sutton1999between}
Sutton, R.~S., Precup, D., and Singh, S.
\newblock Between {MDP}s and semi-{MDP}s: A framework for temporal abstraction
  in reinforcement learning.
\newblock \emph{Artificial intelligence}, 112\penalty0 (1-2):\penalty0
  181--211, 1999.

\bibitem[Tessler et~al.(2017)Tessler, Givony, Zahavy, Mankowitz, and
  Mannor]{tessler2017deep}
Tessler, C., Givony, S., Zahavy, T., Mankowitz, D.~J., and Mannor, S.
\newblock A deep hierarchical approach to lifelong learning in minecraft.
\newblock In \emph{AAAI Conference on Artificial Intelligence (AAAI)},
  volume~3, pp.\ ~6, 2017.

\bibitem[Todorov(2007)]{todorov2006linearly}
Todorov, E.
\newblock Linearly-solvable {M}arkov decision problems.
\newblock In \emph{Advances in Neural Information Processing Systems (NIPS)},
  pp.\  1369--1376. MIT Press, 2007.

\bibitem[Toussaint(2009)]{toussaint2009robot}
Toussaint, M.
\newblock Robot trajectory optimization using approximate inference.
\newblock In \emph{International Conference on Machine Learning (ICML)}, pp.\
  1049--1056. ACM, 2009.

\bibitem[Vezhnevets et~al.(2017)Vezhnevets, Osindero, Schaul, Heess, Jaderberg,
  Silver, and Kavukcuoglu]{vezhnevets2017feudal}
Vezhnevets, A.~S., Osindero, S., Schaul, T., Heess, N., Jaderberg, M., Silver,
  D., and Kavukcuoglu, K.
\newblock Feudal networks for hierarchical reinforcement learning.
\newblock \emph{arXiv preprint arXiv:1703.01161}, 2017.

\bibitem[Ziebart(2010)]{ziebart2010modeling}
Ziebart, B.~D.
\newblock \emph{Modeling purposeful adaptive behavior with the principle of
  maximum causal entropy}.
\newblock PhD thesis, 2010.

\bibitem[Ziebart et~al.(2008)Ziebart, Maas, Bagnell, and
  Dey]{ziebart2008maximum}
Ziebart, B.~D., Maas, A.~L., Bagnell, J.~A., and Dey, A.~K.
\newblock Maximum entropy inverse reinforcement learning.
\newblock In \emph{AAAI Conference on Artificial Intelligence (AAAI)}, pp.\
  1433--1438, 2008.

\end{thebibliography}
\bibliographystyle{icml2018}

\FloatBarrier
\newpage

\appendix
\section{Hyperparameters}
\subsection{Common Parameters}
We use the following parameters for LSP policies throughout the experiments. The algorithm uses a replay pool of one million samples, and the training is delayed until at least 1000 samples have been collected to the pool. Each training iteration consists of 1000 environments time steps, and all the networks (value functions, policy scale/translation, and observation embedding network) are trained at every time step. Every training batch has a size of 128. The value function networks and the embedding network are all neural networks comprised of two hidden layers, with 128 ReLU units at each hidden layer. The dimension of the observation embedding is equal to two times the number of action dimensions. The scale and translation neural networks used in the real NVP bijector both have one hidden layer consisting of number of ReLU units equal to the number of action dimensions. All the network parameters are updated using Adam optimizer with learning rate $3\cdot 10^{-4}$.

\autoref{table:common-benchmark-params} lists the common parameters used for the LSP-policy, and \autoref{table:env-benchmark-params} lists the parameters that varied across the environments.

\begin{table}[tbh]
    \caption{Shared parameters for benchmark tasks}
  \begin{center}
    \begin{tabular}{lr}
      \toprule
      Parameter & Value  \\
      \midrule
      learning rate                                   & $3\cdot 10^{-4}$      \\
      batch size                                      & 128      \\
      discount                                        & 0.99      \\
      target smoothing coefficient                    & $10^{-2}$      \\
      maximum path length                             & $10^{3}$      \\
      replay pool size                                & $10^{6}$       \\
      hidden layers (Q, V, embedding)                & 2       \\
      hidden units per layer (Q, V, embedding)       & 128       \\
      policy coupling layers                          & 2       \\
      \bottomrule
    \end{tabular}
  \end{center}
  \label{table:common-benchmark-params}
\end{table}

\begin{table}[tbh]
  \caption{Environment specific parameters for benchmark tasks}
  \centering
  \begin{tabular}{l|rrrrrr}
    \toprule
    Parameter                 
    & \rotatebox[origin=c]{90}{Swimmer (rllab)}
    & \rotatebox[origin=c]{90}{Hopper-v1}
    & \rotatebox[origin=c]{90}{Walker2d-v1}
    & \rotatebox[origin=c]{90}{HalfCheetah-v1}    
    & \rotatebox[origin=c]{90}{Ant (rllab)}       
    & \rotatebox[origin=c]{90}{Humanoid (rllab)} \\
    \midrule
    action dimensions            & 2 & 3 & 6 & 6 & 8 & 21 \\
    reward scale                & 100 & 1 & 3 & 1 & 3 & 3 \\
    observation embedding dimension & 4 & 6 & 12 & 12 & 16 & 42 \\
    scale/translation hidden units & 2 & 3 & 6 & 6 & 8 & 21 \\
    \bottomrule
  \end{tabular}
  \label{table:env-benchmark-params}
\end{table}

\subsection{High-Level Policies}
All the low-level policies in hierarchical cases (Figures \ref{fig:ant_hierarchy}, \ref{fig:benchmark_stack_ant}, \ref{fig:benchmark_stack_humanoid}) are trained using the same parameters used for the corresponding benchmark environment. All the high-level policies use Gaussian action prior. For the Ant maze task, the latent sample of the high-level policy is sampled once in the beginning of the rollout and kept fixed until the next one. The same high-level action is kept fixed over three environment steps. Otherwise, all the policy parameters for the high-level policies are equal to the benchmark parameters.

The environments used for training the low-level policies are otherwise equal to the benchmark environments, except for their reward function, which is modified to yield velocity based reward in any direction on the xy-plane, in contrast to just positive x-direction in the benchmark tasks. In the Ant maze environment, the agent receives a reward of 1000 upon reaching the goal and 0 otherwise. In particular, no velocity reward nor any control costs are awarded to the agent. The environment terminates after the agent reaches the goal.

\end{document}